\documentclass[]{spie}  

\usepackage{amsmath,amsfonts,amssymb}
\usepackage{graphicx}
\usepackage{subfig}
\usepackage{setspace}
\usepackage{booktabs}
\usepackage{placeins}
\usepackage{float}

\title{Vendor-independent soft tissue lesion detection using weakly supervised and unsupervised adversarial domain adaptation}

\author[b]{Joris van Vugt}
\author[b]{Elena Marchiori}
\author[a]{Ritse Mann}
\author[c]{Albert Gubern-M\'erida}
\author[a]{Nikita Moriakov}
\author[a,d]{Jonas Teuwen}

\affil[a]{Radboud University Medical Center, Diagnostic Image Analysis Group, Department of Radiology and Nuclear Medicine, Nijmegen, the Netherlands}
\affil[b]{Radboud University, Nijmegen, the Netherlands}
\affil[c]{Screenpoint Medical, Nijmegen, the Netherlands}
\affil[d]{Optics Research Group, Imaging Physics Department, Delft University of Technology, the Netherlands}

\authorinfo{Send correspondence to Jonas Teuwen: jonas.teuwen@radboudumc.nl.}

\begin{document} 
\maketitle

\begin{abstract}
Computer-aided detection aims to improve breast cancer screening programs by helping radiologists to evaluate digital mammography (DM) exams. DM exams are generated by devices from different vendors, with diverse characteristics between and even within vendors. Physical properties of these devices and postprocessing of the images can greatly influence the resulting mammogram. This results in the fact that a deep learning model trained on data from one vendor cannot readily be applied to data from another vendor. This paper investigates the use of tailored transfer learning methods based on adversarial learning to tackle this problem. We consider a database of DM exams (mostly bilateral and two views) generated by Hologic and Siemens vendors. We analyze two transfer learning settings: 1) unsupervised transfer, where Hologic data with soft lesion annotation at pixel level and Siemens unlabelled data are used to annotate images in the latter data; 2) weak supervised transfer, where exam level labels for images from the Siemens mammograph are available. We propose tailored variants of recent state-of-the-art methods for transfer learning which take into account the class imbalance and incorporate knowledge provided by the annotations at exam level. Results of experiments indicate the beneficial effect of transfer learning in both transfer settings. Notably, at  0.02 false positives per image, we achieve a sensitivity of 0.37, compared to 0.30 of a baseline with no transfer. Results indicate that using exam level annotations gives an additional increase in sensitivity.
\end{abstract}

\section{Introduction}
Population-based breast cancer screening programs with mammography have proven to reduce mortality and the morbidity associated with advanced stages of the disease. Radiologists have to evaluate a large amount of mammograms with a very low prevalence of malignant cases in a short period of time, which leads to possible interpretation errors. To alleviate this, computer-aided diagnosis (CAD) systems can be employed. Most current CAD systems are based on deep learning algorithms which work directly on the data and do not require any feature engineering. This however comes with the requirement of a lot of annotated training data which is expensive and time-consuming to acquire. The variability in DM between the different vendors, and between different mammographs of the same vendor further complicates this task. As deep learning algorithms are usually sensitive to this type of variation, this causes the problem that a model trained on mammograms from one vendor cannot readily be applied to mammograms produced by another vendor.

In machine learning this problem is refered to as domain shift and is addressed by transfer learning methods. Perhaps the simplest way to address this is problem is to collect labeled data from each vendor. While very effective this is obviously not very practical. When such labelled data is not available, adversarial methods, which belong to the state of the art transfer learning, can be employed. Here we investigate the use of adversarial transfer learning with no or weak supervision. Specifically, we consider two transfer learning settings: 1) unsupervised transfer, where Hologic data with soft lesion annotations at pixel level and Siemens unlabelled data are used to annotate images in the latter data; and 2) weak supervised transfer, where exam level labels for images from the Siemens mammograph are available. This latter setting is motivated by the observation that exam-level annotations are considerably cheaper to acquire than pixel-level one.

We tailor recent state of the art adversarial transfer learning methods to take into account the skewness of the annotation and to incorporate knowledge provided by annotation at exam level.
Results of our experiment indicate the effectiveness of the proposed methods in both settings.

\section{Methods and materials}
\subsection{Patient population and ground truth labeling}\label{sec:data}
This study was conducted with anonymized data retrospectively collected from our institutional archive. DM exams from women attending the national screening program at our collaborator institution, or our institution for diagnostic purposes between 2000 and 2016 were included.

All malignant lesions were verified by histopathology and manually annotated and delineated under the supervision of an expert breast radiologist. The annotator had access to other breast imaging exams, radiological and the histopathological reports. The normal cases were selected if they had at least two years of negative follow-up. This yielded a total of 5009 DM exams, from which 22\% of the exams contained a total of 1731 biopsy-verified malignant lesions. Most exams were bilateral and included two views (cranio-caudal -CC- and medio-lateral oblique -MLO-).
The images uses in this study were acquired by mammographs from two different vendors which served as the source and target domain respectively (Selenia Dimensions, Hologic, USA; Mammomat Inspiration and Mammomat Novation DR, Siemens, Germany) which have a pixel spacing of $70\mu$m and $85\mu$m respectively.

\subsection{Model architecture and training procedure}

We apply a two stage model. In particular we apply the candidate detector of Karssemeijer\cite{karssemeijer1996detection} which gives about 15 candidates per image at a near 100\% sensitivity. Around these locations, we extract patches and classify these as malignant/benign using a convolutional neural network (CNN) which we describe below. Prior to classification, all images were bilinearly resampled to a $200\mu$m pixel spacing. This resolution is considered a good trade-off between accuracy, memory usage and speed for the detection of soft tissue lesions.

\subsection{Deep learning network architecture and training}
For every experiment in this paper we use a network architecture based on VGG net. We use six blocks of two $3 \times 3$ convolutions with ReLU activation function and batch normalization followed by max pooling with stride $2$. We take 16 initial filters, and double this each convolution layer to end with a convolution layer with $512$ filters. We flatten the final activation maps using a global max pooling layer. The classifier consists out of with two fully connected layers with 256 and 1 unit(s) respectively. We observed no performance difference with other architectures such as ResNet or Densenet.

For each experiment, we train on the same source dataset (Hologic) and apply the following augmentations: random horizontal and vertical flips, rotations up to 15 degrees, zoom in/out up to 10\%, and translation of up to 15 pixels (3mm). We use balanced batches of size 64 and train for 10 epochs. Adam was used as optimizer with a learning rate of 0.001 for the first 6 epochs and 0.0002 for the last 4 epochs, where an epoch is defined as seeing all negative patches once. Each method was trained for 1000 iterations. 

In the subsequent domain adaptation step, where we adapt the model to Siemens mammograms, we apply three recent transfer learning algorithms based on adversarial learning: RevGrad \cite{ganin2014unsupervised}, ADDA \cite{tzeng2017adversarial} and WDGRL \cite{shen2017adversarial}. Motivated by previous work on transfer learning, we apply a class balancing scheme in the domain adaptation set as follows: every 200 iterations, we use the current model to predict the label of the target samples, and we use these \emph{pseudo-labels} to balance the batches. We use stochastic gradient descent (SGD) to train the domain adaptation models.

\subsection{Weak supervision setting}
In addition to the above unsupervised methods, we explore a semi-supervised setting where we have a dataset for which only exam-level labels are available. In particular there means that in at  least one of the four images of the exam, consisting of the CC and and MLO view of both the left and right breast, there is a lesion, but without precise location which is information which is usually readily available. We propose to use these ``weak'' labels to class balance batches with pseudo-labeling. To do this, during training, we use the network to predict the labels for patches only from positive exams. Specifically, per exam only the top four candidates are used as positives, while the rest is discarded. This choice is motivated by the domain knowledge: most lesions have at least 2 candidates and a lesion is visible in both views of that breast. This technique removes a lot of false positives coming from negative cases, at the cost of needing additional information.

\subsection{Evaluation of performance}\label{sec:results}
Performance assessment was computed on a fixed test set of Siemens data split on  patient-level. The performance of the model was evaluated using free receiver operating characteristic (FROC) analysis. The FROC curve is defined as the plot of sensitivity versus the average number of false positives per scan \cite{Moskowitz2017}. In this analysis, a lesion was deemed to have been correctly predicted if the center was within the annotated region.

\section{Results}
\begin{figure}[h!]
    \centering
    \subfloat[]{\includegraphics[width=0.40\linewidth]{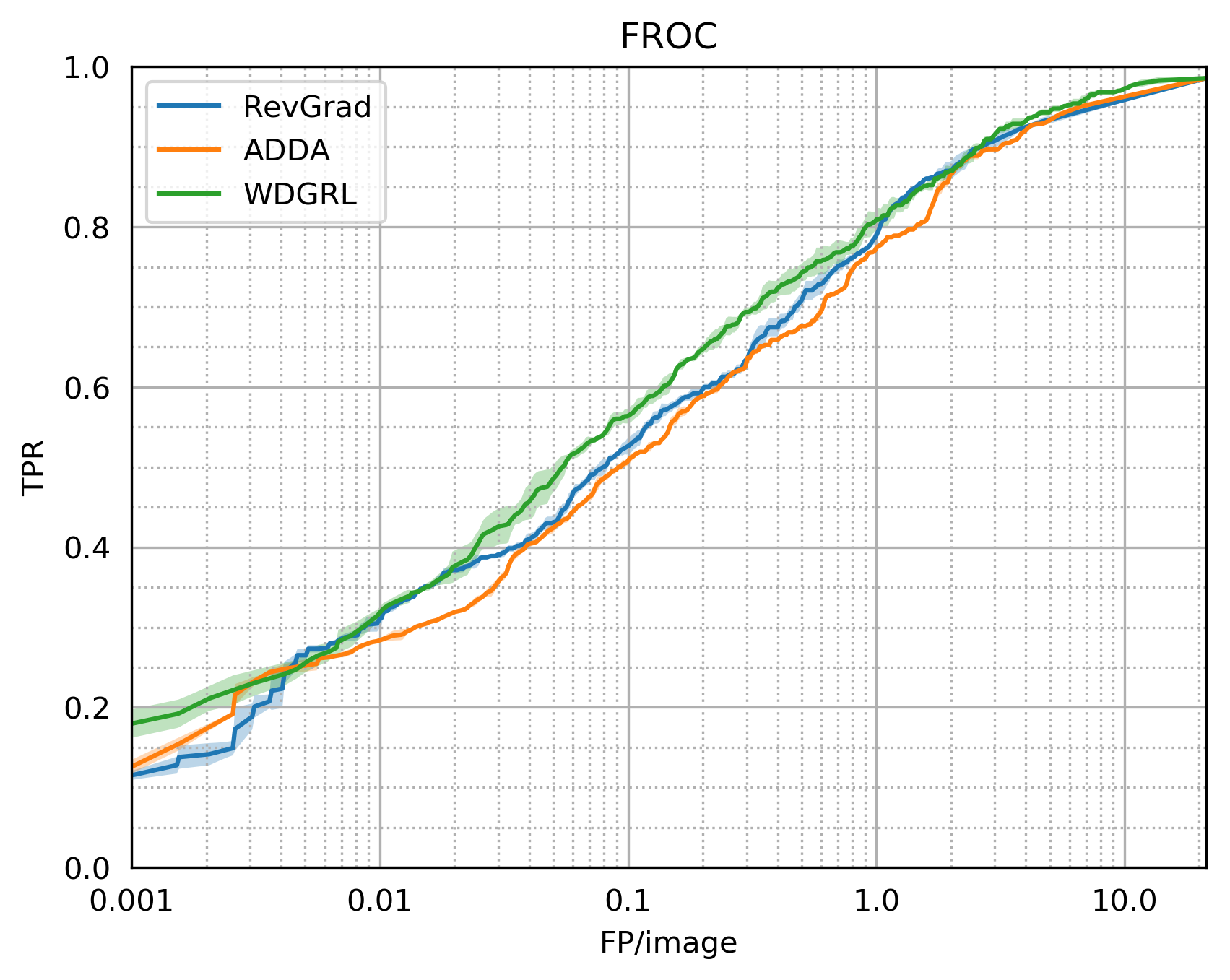}}
    \subfloat[]{\includegraphics[width=0.40\linewidth]{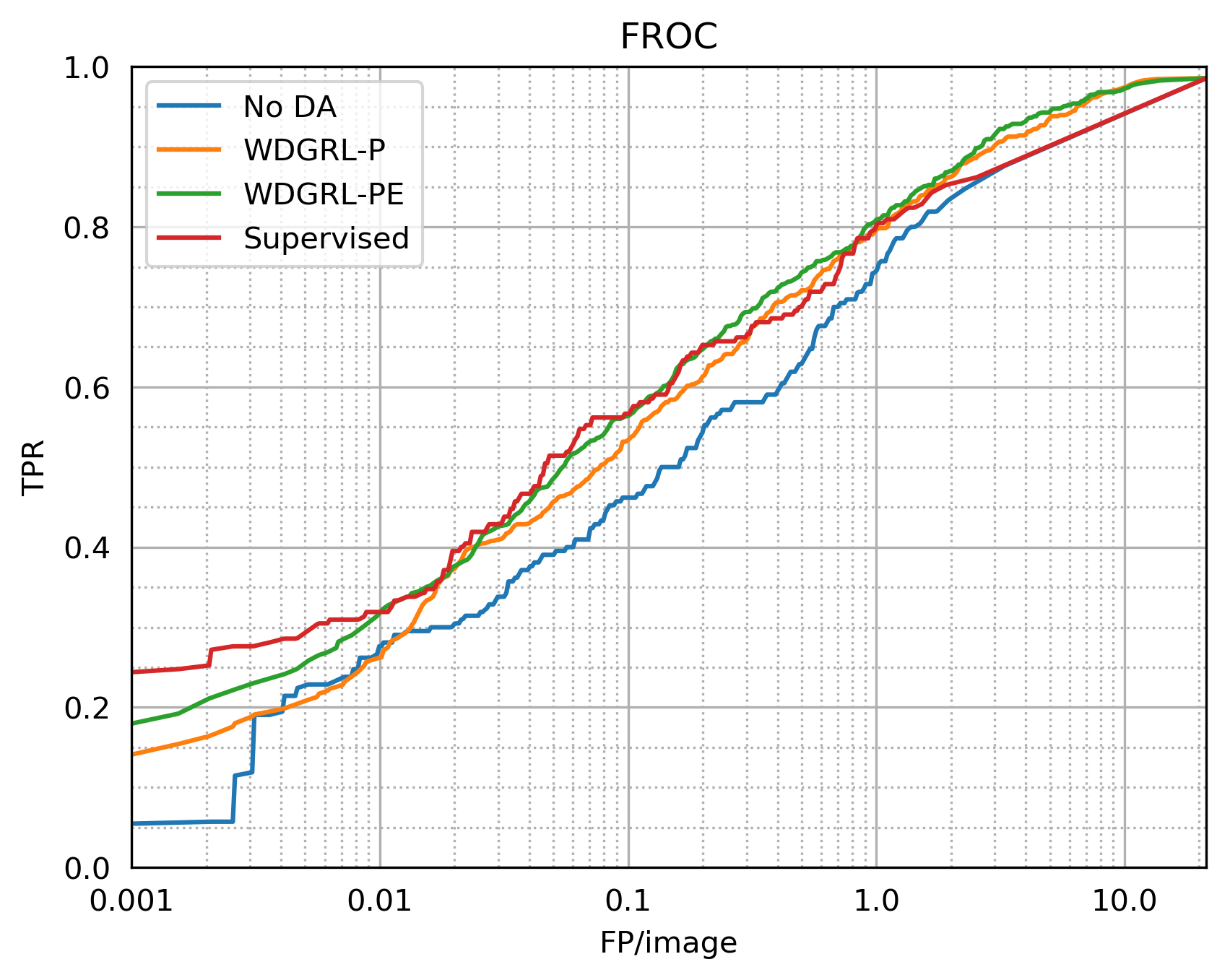}}
    \caption{(a) shows the performance of each algorithm averaged over 3 runs. The shaded regions indicate 1 standard deviation. (b) shows the performance of WDGRL compared to the baseline (no DA) and finetuning the network on the target dataset with supervised learning on a held-out training set (i.e. with labels). In WDGRL-P, pseudolabeling was used to balance the batches. In WDGRL-PE, exam-level labels were incorporated when balancing batches. Supervised is the network that is finetuned on a held-out labeled training set of Siemens data (containing 44 lesions).}
    \label{fig:algorithms}
\end{figure}

Figure \ref{fig:algorithms}a shows the performance of the algorithms on the target domain (Siemens).
Overall, WDGRL (without balancing) outperformed the other two methods and improves substantially over the baseline of directly applying the trained network to the new domain.
It performs similar to a network that is finetuned on a held-out labeled training set of Siemens data (containing 44 lesions).
The performance of the network trained using supervised learning is probably limited by the relatively small size of the training dataset, but this proves that DA can be very effective.
All methods have little variance across runs.
\begin{table}[h!]
\small
    \centering
    \begin{tabular}{lrrr}
        \toprule
        & \multicolumn{3}{c}{Sensitivity} \\
        \cmidrule(r){2-4}
        Method & 0.01 FP/image & 0.02 FP/image & 0.1 FP/image \\
        \midrule
        No domain adaptation & 0.28 & 0.30 & 0.46 \\
        WDGRL (no balancing) & 0.28 & 0.32 & 0.48 \\
        WDGRL-P & 0.26 & 0.37 & 0.53 \\
        RevGrad-PE & 0.31 & 0.37 & 0.53 \\
        ADDA-PE & 0.28 & 0.32 & 0.51 \\
        WDGRL-PE & \textbf{0.32} & \textbf{0.38} & \textbf{0.56} \\
        \hline
        Supervised & 0.32 & 0.40 & 0.57 \\
        \bottomrule
    \end{tabular}
    \caption{Sensitivity at various common false positive levels. P indicates batches are balanced using pseudolabeling. E indicates exam-level labels were used to achieve balanced batches.}
    \label{tbl:results}
\end{table}
Table \ref{tbl:results} reports sensitivity at various common false positive levels. WDGRL-PE achieves best performance among the considered domain adaptation settings and algorithms. Notably, at $0.01$ false positive level, its performance is equal to that of the fully supervised method (Supervised) where we fine-tune a network on a held-out labeled training set of Siemens data.

\section{Conclusion}
We investigated transfer learning in the context of 
soft tissue lesion detection in digital mammography data from different vendors. 
Results of our experiments show that transfer learning can substantially improve the performance of models for soft tissue lesion detection in digital mammography trained on images from another vendor. In the context of lesion detection in mammography, results indicated it might be worth the effort to collect exam-level labels -- which are cheaper to acquire than pixel-level labels -- to get an additional increase in performance. In our approach, this information was only used to class balance the batches during training. It is interesting to investigate in future work a more aggressive exploitation of exam-level labels for improving soft tissue lesion detection at pixel level.

An issue with the transfer learning algorithms used in this paper is their lack of robustness to hyperparameter settings and optimizers. For instance, using the Adam optimizer with WDGRL significantly increases the variance between runs, whereas SGD consistently yields good results.
Furthermore, ADDA and WDGRL were both very sensitive to the ratio in which the feature extractor and discriminator were trained.
If one of these networks became a lot better than the other, the parameters would not converge.
It is claimed that this problem is alleviated by using the Wasserstein distance, but we could not reproduce this type of result in our experiments. We did notice that all adversarial methods used here were robust to the architecture of the domain discriminator.
In our context, we noticed that a one-layer network (i.e., logistic regression) would achieve similar performance to networks with three layers with 400 times as many parameters.

An interesting future research direction is being able to apply a model to images at the native resolution.
Currently, images are downsampled by a factor of approximately three (depending on the vendor) which makes spiculated masses harder to detect. Neural networks have trouble learning discriminative information in larger images and require even larger datasets. Moreover, using domain adaptation to transfer knowledge to new vendors would then also need to take into account the resolution differences between these vendors.

This work has not be submitted for consideration elsewhere.

\small
\bibliographystyle{splncs04}
\bibliography{references}
\end{document}